%% file: root.tex
\documentclass[letterpaper, 10 pt, conference]{ieeeconf}  

\IEEEoverridecommandlockouts                              

\usepackage{graphicx}
\usepackage{bm,microtype}
\usepackage{hyperref}
\usepackage{amsfonts}
\usepackage{xcolor}
\usepackage{cite}

\title{\LARGE \bf FishGym: A High-Performance Physics-based Simulation Framework for Underwater Robot Learning}

\author{Wenji Liu$^{1}$, Kai Bai$^{1}$, Xuming He$^{1}$, Shuran Song$^{2}$, Changxi Zheng$^{2}$ and Xiaopei Liu$^{1}$
 \thanks{$^{1}$ ShanghaiTech University, $^{2}$ Columbia University}
}

\begin{document}

\maketitle
\thispagestyle{empty}
\pagestyle{empty}

\input{abstract.tex}

\input{introduction.tex}

\input{related_work.tex}

\input{main_body.tex}
\input{evaluation_and_conclusion.tex}

\bibliographystyle{IEEEtran}
\bibliography{root}
\end{document}

%% file: abstract.tex
\begin{abstract}

Bionic underwater robots have demonstrated their superiority in many applications. 
Yet, training their intelligence for a variety of tasks that mimic the behavior of underwater creatures poses a number of challenges in practice, mainly due to lack of a large amount of available training data as well as the high cost in real physical environment. 
Alternatively, simulation has been considered as a viable and important tool for acquiring datasets in different environments, but it mostly targeted rigid and soft body systems.
There is currently dearth of work for more complex fluid systems interacting with immersed solids that can be efficiently and accurately simulated for robot training purposes. 
In this paper, we propose a new platform called ``FishGym'', which can be used to train fish-like underwater robots.
The framework consists of a robotic fish modeling module using articulated body with skinning, a GPU-based high-performance localized two-way coupled fluid-structure interaction simulation module that handles both finite and infinitely large domains, as well as a reinforcement learning module. 
We leveraged existing training methods with adaptations to underwater fish-like robots and obtained learned control policies for multiple benchmark tasks. The training results are demonstrated with reasonable motion trajectories, with comparisons and analyses to empirical models as well as known real fish swimming behaviors to highlight the advantages of the proposed platform.   


\end{abstract}

%% file: introduction.tex
\section{Introduction}
\label{sec:intro}

Bio-inspired underwater robots often demonstrate strong maneuverability, propulsion efficiency, and deceptive visual appearance.
These advantages have motivated a set of academic studies on bio-inspired soft robots and biomimetric fish-like robots in the past years~\cite{Du2015,Paley2021,Duraisamy2019}.
It also opens up some important applications, such as marine education, navigation and rescue, seabed exploration, scientific surveying, etc.~\cite{Kopman2012,Picardi2020,Berlinger2021,Li2021,Katzschmann2018}.
However, due to lack of sufficient datasets and high physical cost, training their intelligent behaviors in real environments that at least mimic the bionic creatures or even exceed their capabilities in accomplishing complex tasks is still quite challenging.

Alternatively, simulation has been considered as a viable and important tool for acquiring a large number of datasets in different scenarios~\cite{AvilaBelbutePeres2018,Lee2019,James2019,Bergamin2019}.
Most of the currently available simulators for robot training mainly target rigid and soft body systems~\cite{Coumans2015,Lee2018,Todorov2012}.
Existing simulators for fluid environment are either highly inaccurate (e.g., based on an empirical model~\cite{Terzopoulos1994}), too restrictive to support different agents or  environments~\cite{Song2017,Song2020,Verma2018}, or expensive to generate a large amount of training data \cite{Verma2018,Novati2017}.
There is currently a dearth of simulation platform which is able to provide a versatile, efficient yet accurate results that could be used for training control policies of underwater robots. 

\begin{figure}[t]
	\centering
	\includegraphics[width=0.98\linewidth]{"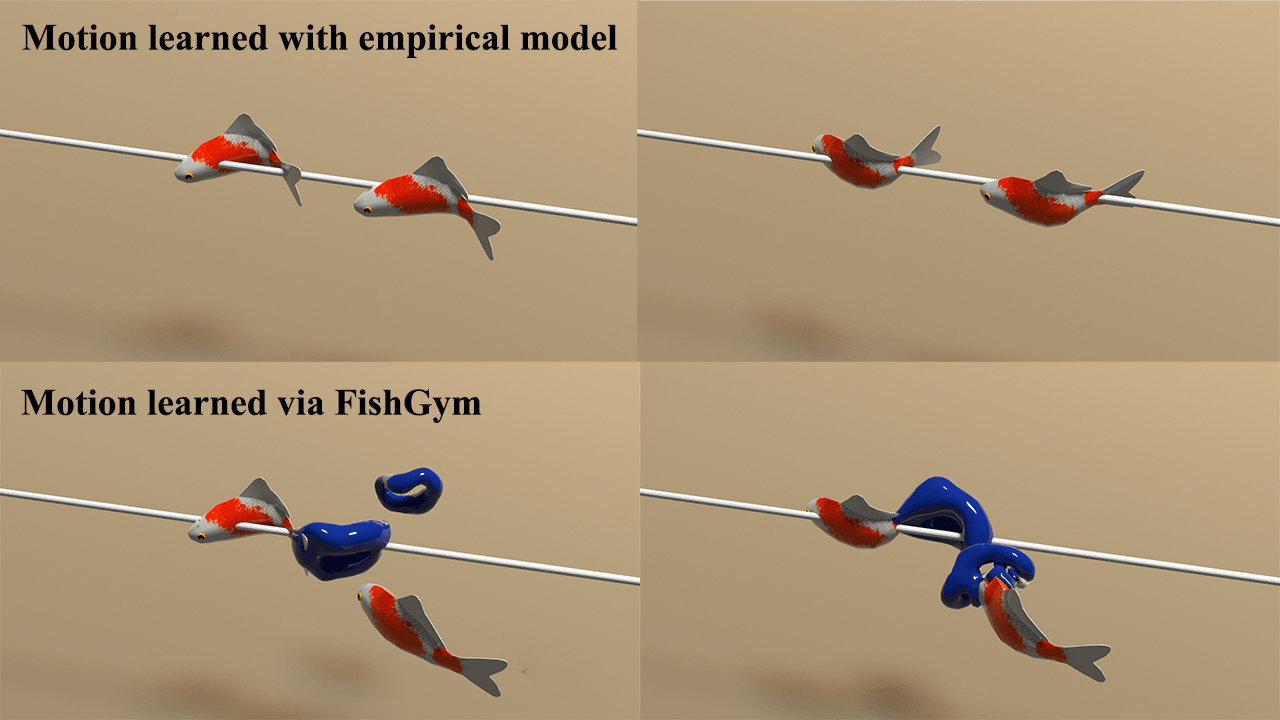"} \vspace{-3mm}
	\caption{\textbf{Two-fish schooling behaviors.}  With the empirical model (1st row), the follower fish always stays in the same line as the leader fish. However, with our physical simulator (2nd row), the follower fish can utilize the wake vortex ring (blue) and gradually pass through the vortex ring to preserve energy. This motion behavior cannot be acquired through a simple empirical model, highlighting the advantages of the proposed FishGym platform. }  \vspace{-3mm}
	\label{fig:teaser}
	\vspace{-10pt}
\end{figure}
We propose \textbf{FishGym}\footnote{https://github.com/fish-gym/gym-fish}, a high-performance simulation platform targeting the two-way interaction dynamics between fish-like underwater robots and the surrounding fluid environment. The robots are modeled by the skeletons of arbitrary topology with surface skinning, whose motion is driven by the articulated rigid body dynamics~\cite{Weinstein2006}, while the fluid-structure interaction is achieved using a recently proposed GPU-optimized lattice Boltzmann solver~\cite{Li2020,Chen2021gpu}, where immersed boundary method~\cite{li2016immersed,wu2010improved} is employed for efficient two-way coupling.
To support simulations in a local fluid domain around the robot to enable training in an infinitely large physical domain, we propose a modification of the original lattice Boltzmann solver that enables simulations in a local frame of reference with acceleration, with higher flexibility in acquiring various training environments. 
The whole simulation module is then coupled with a reinforcement learning module implemented using PyTorch~\cite{Raffin2019,Paszke2019}.   

To demonstrate the capability of the proposed simulator, we evaluated existing reinforcement learning algorithm with reward functions and training procedures tailored for underwater robots.
We compare the learned control policies with that from the empirical model on several underwater planning and control tasks to assess the feasibility and advantages of our framework. 
Analyses on the emerged behaviors also indicate consistency with previous studies on fish motion in nature.   
In summary, we have made the following contributions to training bionic underwater robots: 
\begin{itemize}
\item A GPU-accelerated lattice Boltzmann solver that enables high-performance fluid-structure interaction in a local moving frame of reference to allow robot swimming in an infinitely large domain;
\item A high-performance simulation platform to help explore training bionic underwater robots;
\item A learning algorithm tailored for bionic under-water robots that is able to acquire natural and efficient control policies for swimming;
\item A collection of benchmark tasks for underwater robot to evaluate and compare different learning methods and control policies. 

\end{itemize}

%% file: related_work.tex
\section{Related Work}
We herein review the relevant work in the literature for both simulation and learning algorithms before we dig into the details of our whole framework.

\subsection{Simulation environments in robotics}

Robot training can be achieved following OpenAI Gym~\cite{Brockman2016}, which is an open-source robot learning framework with general definitions, and can be implemented for training a variety of robots with different environments.
However, both the simulator and learning framework should be provided separately.
At present, the most commonly used physics simulators in robot are based on rigid-body, soft-body and cloth dynamics~\cite{Coumans2015,Todorov2012}. 
In particular, for articulated rigid body systems, DART~\cite{Lee2018} can be a good choice.

Fluid environment was traditionally provided by solving N\"avier-Stoke (NS) equation coupled with a rigid body simulator~\cite{Tan2011}; but efficiency limited their application especially for vortical flows. Recently, a new simulation environment for underwater soft-body creatures appeared relying on the finite-element method and projection dynamics~\cite{min2019softcon}; however, its choice of empirical formula~\cite{Terzopoulos1994} on hydrodynamics makes it impossible to create complex flow environment involving vortices and turbulence. The same issue also applies to some marine vehicle simulators~\cite{cieslak2019stonefish,manhaes2016uuv} based on Fossen model\cite{fossen2011handbook}.
Very recently, Gan et al.~\cite{Gan2020} proposed a fluid environment, but only for limited tasks and accuracy.

There is lack of versatile and efficient yet accurate fluid simulation environment upon which more general underwater robot training can be performed, and our proposed ``FishGym'' tries to fill the gap by providing highly efficient GPU-based simulator for two-way coupled fluid-structure interaction.
There are also some learning frameworks that can be used based on OpenAI Gym, e.g., rllib~\cite{Liang2018}, Coach~\cite{Caspi2017} and stable-baselines3~\cite{Raffin2019}, and we adopted ``stable-baselines3'' for training our fish-like underwater robots.

\subsection{Fluid-structure interaction}

Fluid simulation has been studied for decades.
Two different fields have intensively progressed its development.
In computational fluid dynamics (CFD), fluid simulation mostly targets accuracy, and a set of fluid solvers are available, from finite difference \cite{godunov1959finite,Rai-1991,smolarkiewicz1998mpdata}, to finite volume \cite{eymard2000finite,versteeg2007introduction,pinelli2010immersed}, as well as to finite elements \cite{wilson1983finite,girault2012finite,elman2014finite}.
These algorithms are typically very expensive, which are difficult for training underwater robots.
In computer graphics (CG), fluid simulation concerns efficiency more than accuracy, and a large number of more efficient yet less accurate solvers were proposed~\cite{Stam2001,Kim2005FlowFixer,Becker2007Weakly,Ihmsen2014,Jiang2015Affine,Zehnder2018,Qu2019}.
However, even though GPU acceleration has been used in some of these solvers, efficiency is still not high enough.
When rigid body dynamics is coupled for fluid-structure interaction \cite{klingner2006fluid,lv2010novel,dai2005adaptive}, the efficiency can be even lower.

In recent years, lattice Boltzmann method (LBM) has been considered as a very promising alternative to traditional fluid solvers~\cite{Liu-2012,Daniel-2014,Rosis-2017,Li-2018,Li-2020}, exhibiting excellent efficiency and accuracy (usually an order of magnitude faster than the NS counterpart with comparable accuracy on GPU).
Its pure local dynamics without solving global equations greatly benefits the highly parallel implementation~\cite{Li-2018,Li-2020,Chen2021gpu}.
When LBM is coupled with immersed boundary (IB) method~\cite{Li-2020}, it can be easily used to simulate two-way coupled fluid-structure interaction.
In particular, Chen et al.~\cite{Chen2021gpu} proposed a GPU-optimized implementation of IB-LBM, which provides a super-efficient solver for fluid-structure interaction, making the originally expensive fluid simulation now affordable for robot training purposes.
Our proposed platform is based on such a solver, with modifications to allow it for simulating fish dynamics in a local moving domain for higher flexibility, which is not supported in any previous works.

\subsection{Reinforcement learning for robot control}
Reinfocement learning (RL) is a branch of machine learning which aims to train agents using data collected through interaction with the surrounding environment. 
For real world problems in robotics, model-free RL algorithms are often used~\cite{Nian2020}. 
There are two main approaches of model-free RL: policy optimization and Q-learning. 
Policy optimization algorithms, like PPO~\cite{Schulman2017a} and A2C/A3C~\cite{Mnih2016}, are stable but sample-inefficient. 
Q-learning methods, like DQN~\cite{Mnih2013} and C51~\cite{Bellemare2017}, are more sample-efficient but less stable.
Both of them have wide applications. 
For example, PPO was used in multi-robot collision avoidance task~\cite{long2018towards}, bipedal robot locomotion~\cite{li2021reinforcement} etc. 
DQN also proves to work well on a certain type of tasks in real robots~\cite{kato2017autonomous,xin2017application,chen2021non}.
Recently, SAC~\cite{Haarnoja2018} emerges to combine the strengths of the above two main approaches and has proved its capability in real robot problems like Dexterous manipulation~\cite{haarnoja2018soft}, mobile robot navigation~\cite{de2021soft}, robot arm control~\cite{wong2021motion}, multi-legged robot~\cite{haarnoja2018learning}, etc.
Due to its sample efficiency and wide applications, we adopt SAC in this paper.

%% file: main_body.tex
\section{FishGym Framework}
Our physics-based robot learning framework  consists of three components:
1) the robot model for which we focus on fish-like robots, 2) 
the simulation method for predicting fluid-robot interaction, 
and 3) the robot learning method that leverages our simulation method.
We now present their details.


\begin{figure}[t]
	\centering
	\includegraphics[width=0.48\textwidth]{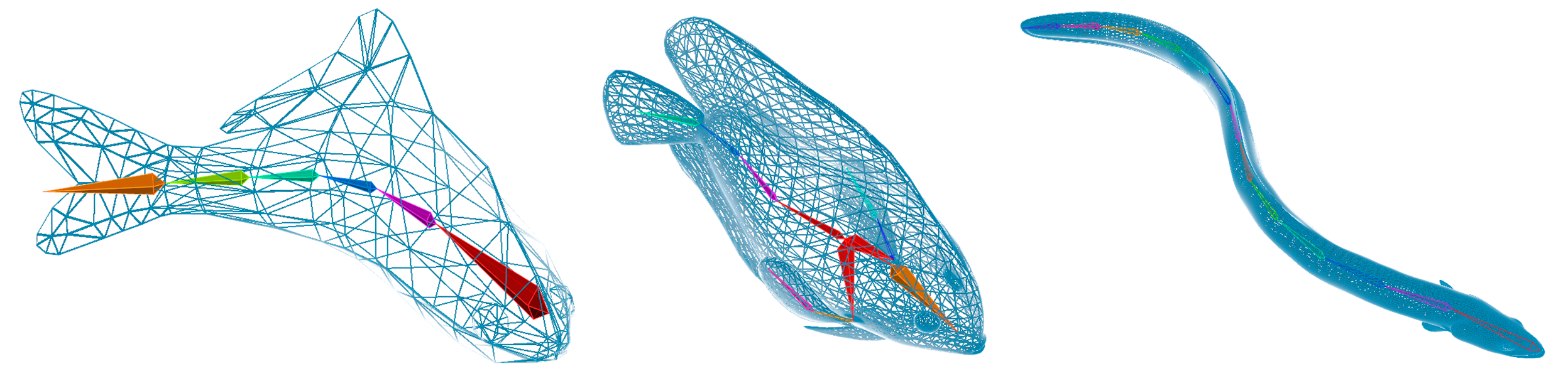}
	\caption[Fish Skeletons]{Illustration of modeled fishes with different skeletons and skins, where the number of joints, the length of each skeleton edge as well as the topology of the skeleton can vary for different types of fishes.}
	\label{fig:fish_model} 
	\vspace{-20pt}
\end{figure}

\subsection{Robot model}
Motivated by the anatomy of the fish structure,
we model a fish robot's locomotion by its skeletal structure, that is, bones connected by joints.
Covering the skeleton are flesh and skin. Provided a skeleton configuration (e.g., with a certain set of joint angles), we use linear blend skinning~\cite{magnenat1988joint} to determine 
the fish's surface shape. In our fluid-robot simulation, the fish surface is assumed to be inelastic, and the flesh is treated as a rigid body under the current skeleton configuration. We make this assumption for the sake of computational efficiency.
By changing its joint angles, a fish robot can adjust its skeleton pose,
which in turn determines its surface shape.
Three examples of fish robots with different skeletal structures are shown in Fig.~\ref{fig:fish_model}.



The bone skeleton is driven by the articulated rigid body dynamics \cite{Weinstein2006}:
\vspace{-8pt}
\begin{equation}
\mathbf{M}(\mathbf{q})\ddot{\mathbf{q}}+\mathbf{C}(\mathbf{q}, \dot{\mathbf{q}})=\boldsymbol{\tau}_{int}+\boldsymbol{\tau}_{ext} ,
\vspace{-8pt}
\end{equation}
where $\mathbf{q}, \dot{\mathbf{q}}$ and $\ddot{\mathbf{q}}$ are respectively the vectors of generalized positions, velocities and accelerations of all joints. $\mathbf{M}(\mathbf{q})$ is the mass matrix, and $\mathbf{C}(\mathbf{q}, \dot{\mathbf{q}})$ accounts for the Coriolis and centrifugal forces;
details of these forces will be described shortly.
$\boldsymbol{\tau}_{ {int }}$ and $\boldsymbol{\tau}_{ {ext }}$ are the vectors representing the generalized internal forces (including the spring forces on joints to enable elasticity, damping forces due to velocities, friction forces, as well as the actuation given by the controller) and the generalized external forces (caused by gravity, possible collisions and the surrounding fluids) exerted on the multi-body system.
We employ DART \cite{Lee2018} to solve the above multi-body system, and for each time step, we control the bone shape by applying generalized actuation forces on joints.
The skin surface is achieved by employing linear blending method proposed by \cite{magnenat1988joint}.

\subsection{GPU-accelerated localized fluid-structure interaction}
There exist many simulation methods that may 
predict fluid-robot interaction~\cite{klingner2006fluid,lv2010novel,dai2005adaptive}. 
These methods, however, require a fixed simulation domain, inside which the 
underwater robot moves. 
When the simulation domain is large,
simulation is costly.
To reduce the cost,
we assume that the fluid 
further away from the robot by a certain distance will not influence the robot motion. 
Thereby, we can fix the size of the simulation domain centered around the fish robot and allow the domain to move along with the fish. 
This setup allows the fish robot to move in an 
infinite spatial domain while keeping the simulation domain limited. 
But then, to capture fluid dynamics correctly, we need to simulate fluid-structure interaction in a moving frame of reference.
Being able to swim in an infinitely large domain is very important for training fish-like underwater robots; Also crucial is the simulation performance, as
training the learning algorithm will often run fluid simulations
many times (see Section~\ref{sec:fish_learning}).
We tackle both problems next.

\subsubsection{Formulation}
Fluid in a fixed frame of reference is often governed by the following NS equation:
\vspace{-6pt}
\begin{equation}
\frac{\partial \mathbf{u}}{\partial t}+(\mathbf{u} \cdot \nabla) \mathbf{u}=-\frac{1}{\rho} \nabla p+\nu \nabla^{2} \mathbf{u} + \mathbf{F},
\vspace{-6pt}
\end{equation}
where $\rho$, $\mathbf{u}$, $p$ and $\mathbf{F}$ represent the density, velocity, pressure and external force fields, and $\nu$ is the kinematic viscosity.
This equation cannot be used to solve flows in a moving frame of reference around the fish, which should be reformulated in a frame of reference with acceleration.
According to \cite{Asmuth2016}, by transforming with time-dependent relative translation $\mathbf{p}$ and rotation $\mathbf{r}$ (represented as Euler angles) between consecutive frames, the NS equation in an accelerating frame of reference results in an additional virtual force added to the system:
\vspace{-6pt}
\begin{equation}
	\vspace{-6pt}
\mathbf{F}_{ni}=-\ddot{\mathbf{p}}-\ddot{\mathbf{r}} \times \mathbf{x}^{\prime}-\dot{\mathbf{r}} \times\left(\dot{\mathbf{r}} \times \mathbf{x}^{\prime}\right)-2 \dot{\mathbf{r}} \times \mathbf{u}^{\prime} ,
\end{equation}
where all the physical quantities are measured in a moving frame of reference.
In case of any immersed solid, e.g., the swimming robot, we apply Neumann boundary condition (i.e., slipping) as an approximation.

\subsubsection{Simulation}
\begin{figure}[t]
	\centering
	\includegraphics[width=0.98\linewidth]{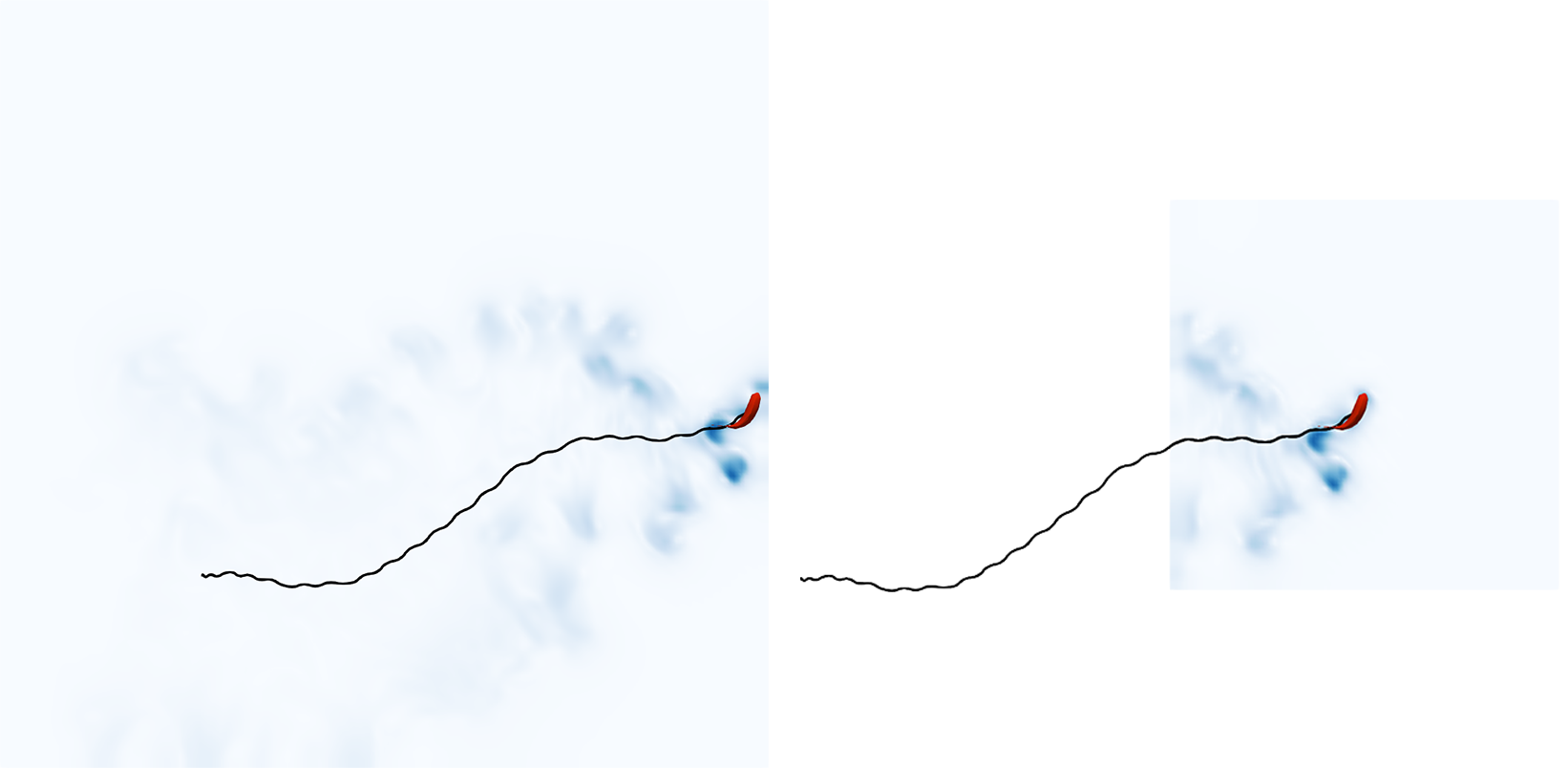} \vspace{-3mm}
	\caption{Comparison between fluid-structure interaction in a global static fluid domain (left) and the non-inertial counterpart in a moving local fluid domain (right), indicating the closeness of the paths with a known policy.} \vspace{-5mm}
	\label{fig:complocalglobal}
	\vspace{-3pt}
\end{figure}

To simulate the above dynamics in an efficient manner, we discard the traditional NS solver; instead, we employ a GPU-optimized LBM solver with immersed boundary (IB) method \cite{Li-2020} for fluid-structure interaction, whose high efficiency has been demonstrated.
The fish-like robot surface is uniformly sampled before simulation, and the external virtual force due to acceleration can be added into the system very easily.
The difficulty we need to address is the domain boundary, which in theory should be set according the large-domain simulation.
However, in practice, we do not know the exact values of the domain boundary, and when fish moves with different velocities and accelerations, the flow can go into and outside from any portion of the domain boundary.
Thus, we need a domain boundary treatment which can adapt to this situation, and through a set of experiments, we found that the method described in \cite{ZhaoLi2002} satisfy this requirement and has been employed in our IB-LB simulation.
Fig.~\ref{fig:complocalglobal} compares the accuracy between a full global fluid domain simulation (left) where we directly use IB-LB method in \cite{Li-2020} and the proposed localized fluid-structure interaction (right). 
Both methods produce nearly identical robot motion trajectories.

\subsection{Reinforcement learning for fish-like robot control}
\label{sec:fish_learning}
Due to complex fish dynamic model, the control of fish-like robot swimming cannot be simply achieved by a model-based controller.
In addition, since the control input is usually high-dimensional and cannot be fully decoupled, the PID controller cannot be used either.
Thus, reinforcement learning (RL) is a common choice to train complex control policies for swimming.
In this paper, we propose four benchmark tasks that will be trained using RL in order to demonstrate the power of the new simulator together with the learning algorithms.
These benchmark tasks are:
\begin{itemize}
	\item \textbf{Cruising.}
	The robot fish tries to swim to reach a given target location that is a distant away from the robot.
	\item \textbf{Pose control.}
	A robot fish tries to control its pose in order to make a U-turn.
	\item \textbf{Two-fish schooling.}
	A robot fish follows a leader fish as closely as possible, where the leader fish is controlled to swim in a straight path.
	\item \textbf{Path following.}
	A robot fish follows a given arbitrary path as closely and efficiently as possible.
\end{itemize}

In RL, an agent learns a policy for a specific task through repeated interaction with the environment.
Given a state $\mathbf{s}_i$, the RL tries to learn a parametric policy $\pi_{\boldsymbol{\theta}}$, which is usually represented by a fully-connected neural network, to produce an action $\mathbf{a}_i$;
The action can be taken by the agent to transit to the next state $\mathbf{s}_{i+1}$, where a reward $r_i$ is evaluated.
The agent iterates transitions until it satisfies one of the exit conditions, e.g., finite time horizon, or success/failure of a given task.
A parametric policy $\pi_{\boldsymbol{\theta}}$ is learned by finding the optimal parameters $\boldsymbol{\theta}^{*}$ that maximize the expected return:
\vspace{-8pt}
\begin{equation}
	\vspace{-6pt}
J(\theta)=\mathbb{E}_{\tau \sim p_{\theta}(\tau)}\left[\sum_{t=0}^{T} \gamma^{t} r_{t}\right] ,
\end{equation}
where $T$ is the maximum number of control time steps, $\gamma$ is the discounting factor, and $\tau$ is the sampled trajectory containing a sequence of states and actions, i.e., $\tau = (s_0, a_0 , s_1, a_1, ..., s_i, a_i, ..., s_t, a_t)$.
The policy learning can be achieved in two different ways.
Depending on the available resources, we can sample the trajectories from one single task, or from multiple tasks to learn a \textit{global policy}.
However, if the variety of tasks is large, it is time consuming especially when the simulator is not fast enough.
On the other hand, if the tasks can be subdivided into small and simple sub-tasks, we can gather all these sub-tasks together and sample from them to learn a \textit{local policy}, which is expected to be much more generalizable given a relatively small training set, and could be affordable for limited resources.
We adopt both approaches to train different tasks.

In the following, we specify in detail the specific designs on how we train these benchmark tasks.
\paragraph{State} 
For all benchmark tasks, we consider the following state variable:
\vspace{-8pt}
$$
\mathbf{s}=(\mathbf{s}_d, \mathbf{s}_p, \mathbf{s}_r, \mathbf{s}_{task}) ,
\vspace{-6pt}
$$
where $\mathbf{s}_d = (\mathbf{q},\dot{\mathbf{q}})$ contains the dynamic states, with $\mathbf{q}$ the vector of generalized joint positions, and $\dot{\mathbf{q}}$ is the vector of generalized joint velocities;
$\mathbf{s}_p=(\mathbf{p},\dot{\mathbf{p}})$ contains translation, where $\mathbf{p}$ is the relative translation vector between simulation time steps, and $\dot{\mathbf{p}}$ is the relative velocity vector;
$\mathbf{s}_r = \mathbf{r} $ contains rotation (Euler angles);
and $\mathbf{s}_{task}$ contains task-specific states. 
In practice, to reduce input dimension, all state variables are expressed in a local coordinate system. 

\paragraph{Action}
For all benchmark tasks, the action can be generally defined as:
\vspace{-10pt}
$$
\mathbf{a} = (\boldsymbol{\sigma},\Delta v) ,
\vspace{-6pt}
$$
where $\boldsymbol{\sigma}$ is the vector containing the actuation forces applied to the joints, and
$\Delta v$ is the change of the bladder's volume inside the robot fish, which controls buoyancy to enable going up and down in a fluid.
\paragraph{Reward}
The reward for training all benchmark tasks can also be written in a general mathematical form as:
\vspace{-4pt}
$$
r = w_pr_p + w_vr_v + w_rr_r + w_er_e + w_{task}r_{task} ,
\vspace{-4pt}
$$
where $r_p= \exp( -\|\mathbf{p}\|_2)$ and $r_v = \|\dot{\mathbf{p}}\|_2$ drive the robot towards its target position and velocity as fast as possible;
$r_r= 1-\|\mathbf{r}\|_2$ drives the robot towards its target rotation (pose);
$r_e =\|\mathbf{\tau}\|^2$ measures the effort exhausted during the swimming;
$r_{task}$ is a task-specific reward, which will be specified later;
and $w_p,w_v,w_r,w_e,w_{task}$ are the corresponding weights for different components in the reward.
The weights for each benchmark task are listed in Table.~\ref{weight_table}.
\begin{table}[t]	
	\caption{Weights used in the reward of each task} 
	\vspace{-10pt}
	\label{weight_table}
	\begin{center}
	\begin{tabular}{c|c|c|c|c|c}
		& $w_v$ & $w_p$ & $w_r$ & $w_e$ & $w_{task}$ \\ \hline
		Cruising           & 1  &0   & 0.2   & 0.5   & 0          \\
		Pose Control       & 0  &0   & 1     & 0     & 0          \\
		Two Fish Schooling & 0  &1   & 0     & 0.1   & 0          \\
		Path Following     & 1  &0   & 0     & 0.5   & 1       \\
		\hline
	\end{tabular}
\end{center}
\vspace{-25pt}
\end{table}
Our proposed four benchmark tasks are trained using either global or local policy learning approaches we described.
\paragraph{Global policy learning}
For cruising, pose control and two-fish schooling tasks, we use global policy learning with a single input task, meaning that we train robot fish separately for each task, where $r_{task}=0$. Fig.~\ref{fig:tasksequence} shows the snapshots of the swimming results, where the first two rows show cruising inside a shallow and a deep fluid; the third row shows the pose control for U-turn, and the fourth row shows the two-fish schooling result, which has not been demonstrated in previous works.

\paragraph{Local policy learning}
\begin{figure}[htb]
	\vspace{-6pt}
	\centering
	\includegraphics[width=0.98\linewidth]{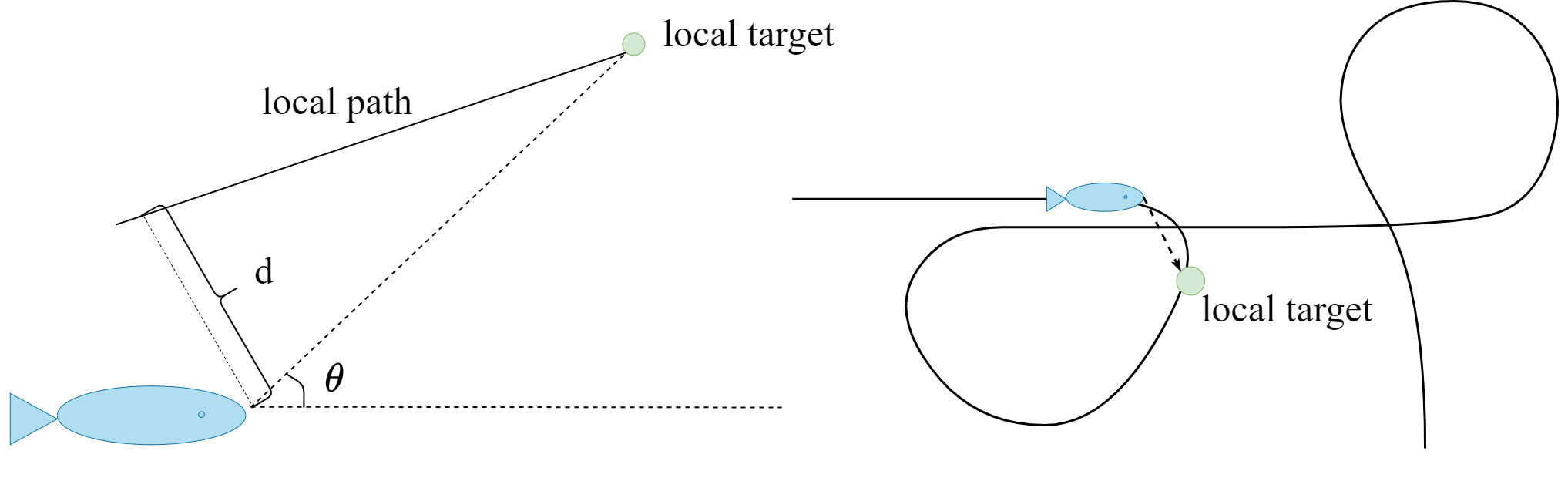}
	\vspace{-12pt}
	\caption{Illustration of local policy training for arbitrary path following. Left: a local target is sampled given a random $d$ and $\theta$; right: when applying learned policy for path following, we always select a local target ahead on the desired path, which changes for each time step.}
	\label{fig:local_policy}
	\vspace{-10pt}
\end{figure}
Training robot fish following an arbitrarily long path is more difficult, and global policy learning could be resource demanding and time consuming.
To make the training easier while also retaining generalizability, we use local policy learning instead.
In fact, following an arbitrarily long path can be viewed as following a sequence of short and straight local paths along nearly the tangent direction of the global path given a robot location, greatly simplifying the training process. 
During training, we randomly sample local paths (parameterized by $(d,\theta)$, see Fig.~\ref{fig:local_policy} (left), where $d$ is the distance to the local path and $\theta$ is the angle to the target; note that we restrict the robot fish to swim in a horizontal 3D plane) and form a set of trajectories that are representative of the local conditions of a global path; then we can train the local policy once and apply it to any specified path at any time step, similar in idea to \cite{Peng2017}. 
In our local policy learning, the task specific state is defined as $s_{task} = \mathbf{d}$, and the task specific reward $r_{task}$ is:
\vspace{-4pt}
\begin{equation}
r_{task} = \|\dot{\mathbf{d}}\|_2+\exp( -\|\mathbf{d}\|_2) ,
\vspace{-4pt}
\end{equation}
which encourages fast and stable convergence to the local path.
Here, $\mathbf{d}$ is a vector containing relative distance to the path.
The training is initialized randomly by the technique proposed in \cite{Peng2018}, and on each trial, random initial velocity is enforced on the robot fish and random angles and velocities are set on the joints.
Once learned, we apply the policy every some time steps based on the input state and a local target location that is 0.5m ahead on the local path, see Fig.~\ref{fig:local_policy} (right), and Fig.~\ref{fig:tasksequence} (bottom) shows a path following result.

%% file: evaluation_and_conclusion.tex
\section{Platform and Evaluation}
\begin{figure}[t]
	\centering
	\includegraphics[width=0.96\linewidth]{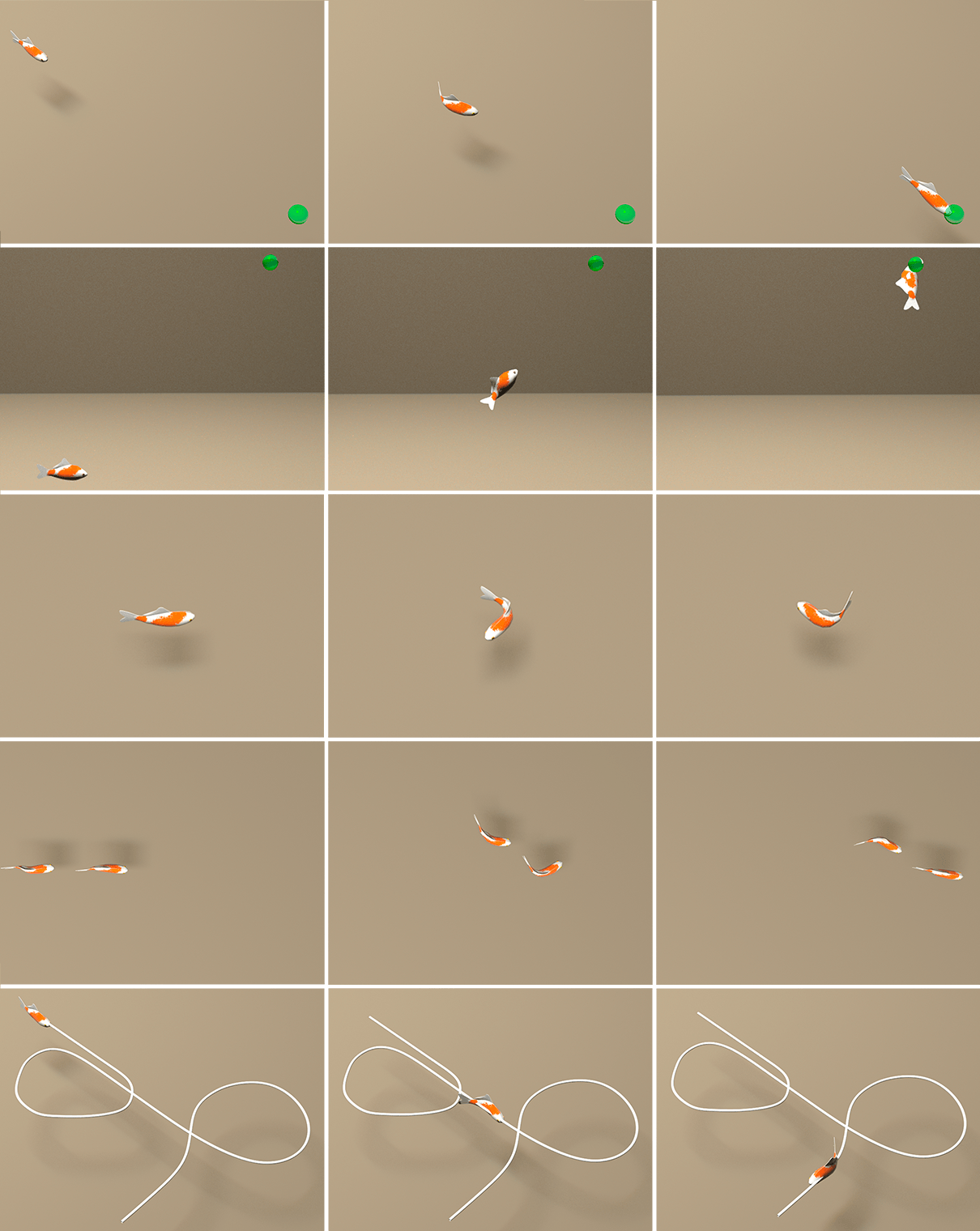} \vspace{-3mm}
	\caption[FishGym Tasks]{The control results for benchmark tasks using our trained policies and two-way coupled fluid-structure interaction solver. Top row: cruising in a shallow fluid; second row: cruising in a deep fluid (with buoyancy control); third row: pose control with a U-turn; fourth row: two-fish schooling; bottom row: path following with an arbitrarily specified path.}
	\label{fig:tasksequence} 
	\vspace{-15pt}
\end{figure}

In the following, we first describe our setup on simulation and evaluate our fluid simulation platform with our learning algorithms for robot fish swimming in multiple aspects.

\subsection{Platform setup}
\paragraph{Simulation}
In most our training tasks, all the fish robots have a density of 1080$kg/m^3$, and we used local non-inertial fluid-structure interaction solver for simulating robot fish dynamics, with a grid resolution of $100\times100\times100$ and a physical time step of 0.004s.
The solver costs around 3.5 seconds for simulating one physical second on an NVidia TitanXp GPU with 12G memory.
For two-fish schooling, we extended the local domain horizontally to simultaneously include two fishes, with a grid resolution of $150\times50\times100$ and the same physical time step, which costs around 4 seconds for simulating one physical second on the same GPU.

\paragraph{Training}
The policy network consists of two layers, each containing 256 units, with an ReLU activation function. 
For each task, the policy is trained using SAC \cite{Haarnoja2018}.
We train each policy for a total of 2000 simulation rollouts, each of which contains 50 time steps, where a candidate policy executes a new action at each time step. 
We train the policy network with a batch size of 256.
The model parameters are updated for each step, and one gradient step is performed after each rollout.
We train all policies on a machine with an NVidia TitanXP GPU, and the training process usually starts to converge after 6 hours (1000 episodes), and have a smooth convergence within 10 hours. 
The training could be several times faster if we use the most state-of-the-art GPUs, such as NVidia GeForce RTX3090.

\subsection{Comparison for different types of robot fishes}
\begin{figure}[t]
	\centering
	\includegraphics[width=0.96\linewidth]{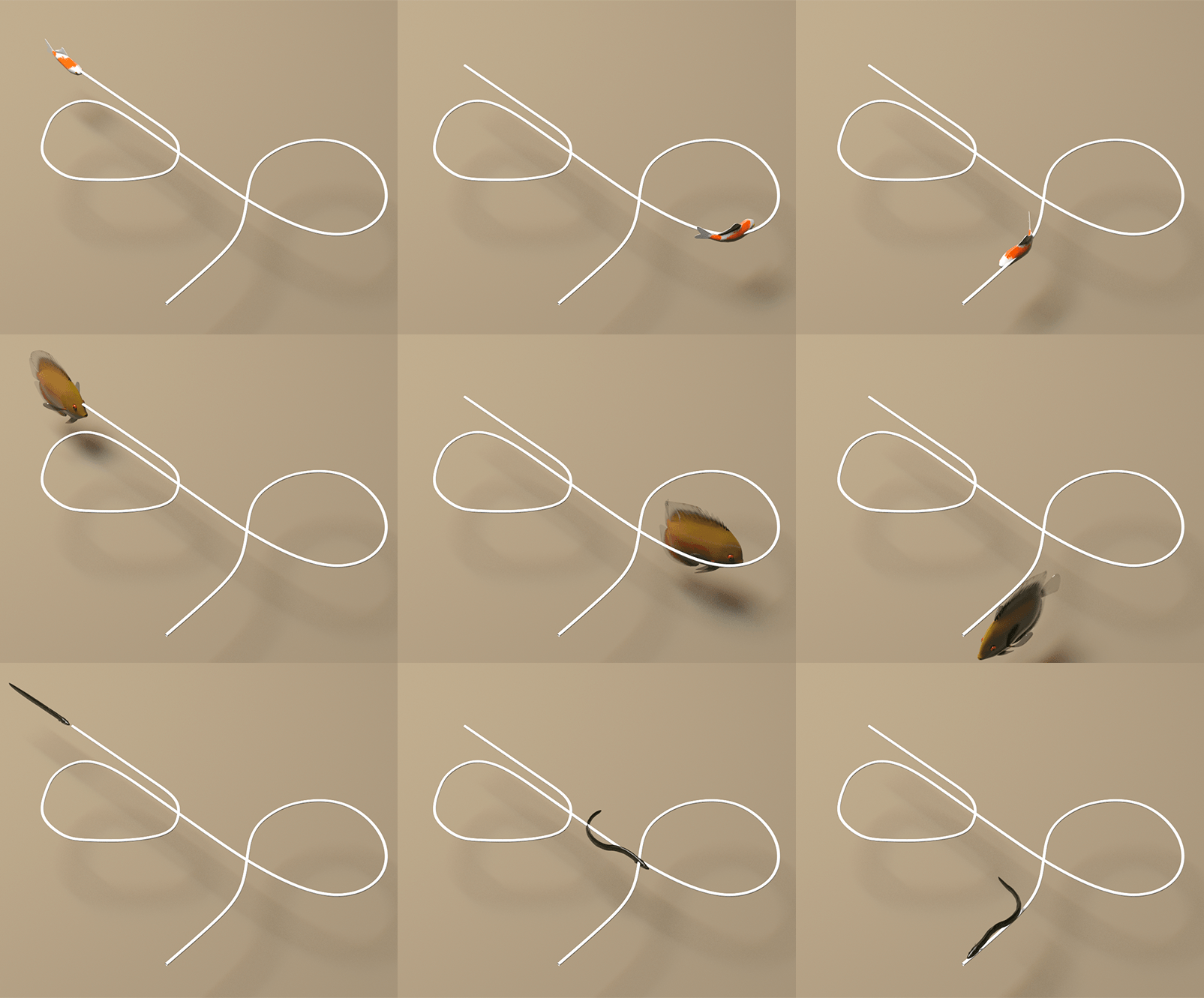} \vspace{-3mm}
	\caption[Fish Gym Tasks]{Different types of robot fishes trained using local policy learning for arbitrary path following. Top: koi robot fish; middle: flatfish robot fish with a different skeleton topology (with branching during modeling); bottom: eel robot fish with a long concatenated skeleton.}\vspace{-3mm}
	\label{fig:different_fish_type_result}
\end{figure}

Our platform can support robot fishes designed with different skeleton connectivity and skin shapes.
Fig.~\ref{fig:different_fish_type_result} shows the training results for three types of robot fishes swimming along an arbitrarily given path.
The average distances from the path (around 7 meters long) are as close as 0.03m, 0.05m, 0.08m, respectively, indicating the capability of our platform in supporting a variety of robot fishes.
\subsection{Comparison for different simulation models}
In the literature, a simple empirical model was proposed as the simulator for robot fish swimming~\cite{Terzopoulos1994}, which has been used until now~\cite{grzeszczuk1998neuroanimator,si2014realistic,min2019softcon}.
It models the instantaneous force on the surface of the robot due to viscous fluid as:
\vspace{-5pt}
\begin{equation}
	F = -k\int_{S}(\mathbf{n} \cdot \mathbf{v}) \mathbf{n} d s ,
	\label{empirical_model_eq}
	\vspace{-5pt}
\end{equation}
where $\mathbf{n}$ is the unit outward normal; $\mathbf{v}$ is the relative velocity between the surface and the fluid (since there is no fluid simulation, the fluid velocity is assumed to be zero), and $k$ is a constant manually tuned for different robot fishes and the surrounding fluids.
Note that for a specific system, $k$ can only be determined either by real measurement data or from other physically more accurate simulators.

To examine the similarity and difference between the empirical model and our physical simulator for robot fish swimming, we conduct two test cases with analyses below.
\paragraph{Path following}
Path following is a primitive task for robot control.
Here, we compare the similarity and difference of path following using an empirical model and our physical simulator.
Since there is no clue on how to tune the parameter $k$ in an empirical model, we arbitrarily choose one and learn a policy. 
\begin{figure}[t]
	\centering
	\includegraphics[width=0.96\linewidth]{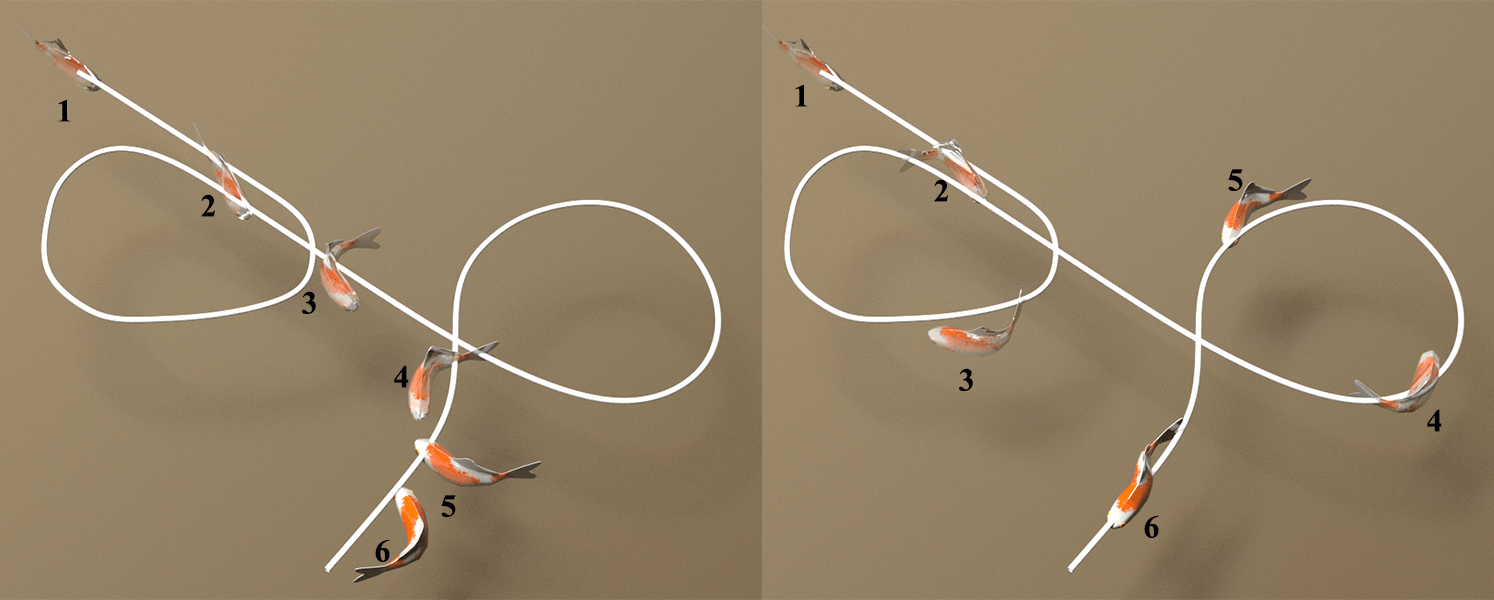} \vspace{-3mm}
	\caption{Comparison for path following using (left) empirical model with an arbitrarily chosen parameter and (right) our physical simulator. Note that due to improper boundary force, serious drifting happens for empirical model when turning. The numbers indicate the order during path following.} \vspace{-3mm}
	\label{fig:empirical_failed_path}
\end{figure}

Fig.~\ref{fig:empirical_failed_path} (left) shows the control result, indicating serious drifting when turning, while our physical simulator does not require any parameter turning (we directly specify the physical parameter for the fluid as $\rho$=1000$kg/m^3$ and $\nu$=0.00089$m^2/s$ with a zero-velocity initialization to match that used in the empirical model), leading to a reasonable path following result as shown in Fig.~\ref{fig:empirical_failed_path} (right).
Note that the mean path deviation for the empirical model with an arbitrary parameter is as large as 0.5m, while our physical simulator only has a mean path deviation of 0.03m.
The empirical model can be much improved if we collect data from our simulator and fit the parameter $k$, leading to a very similar result in a static fluid environment but runs much faster.
However, in some cases where we cannot assume static fluid background, empirical model can completely fail no matter how we fit the parameter $k$, and we demonstrate this case in the following two-fish schooling task.
\paragraph{Two-fish schooling}
Fish schooling describes a common phenomenon where fishes tend to swim in a group and one fish follows the other.
It has been revealed by scientists that due to the vortex ring generated behind the leader fish, the follower fish tries to utilize the vortex ring to reduce drag and pass through it in order to catch up with the leader fish more efficiently \cite{Novati2017,Verma2018}.
We demonstrate this behavior in Fig.~\ref{fig:teaser} (bottom row).
Due to more accurate modeling to capture complex fluid flows, the training can successfully obtain a policy that utilizes the vortex ring, see Fig.~\ref{fig:teaser} (bottom), while empirical model, on the other hand, fails to learn such a policy due to lack of a real fluid-structure interaction.

\begin{figure}[t]
	\centering
	\includegraphics[width=0.98\linewidth]{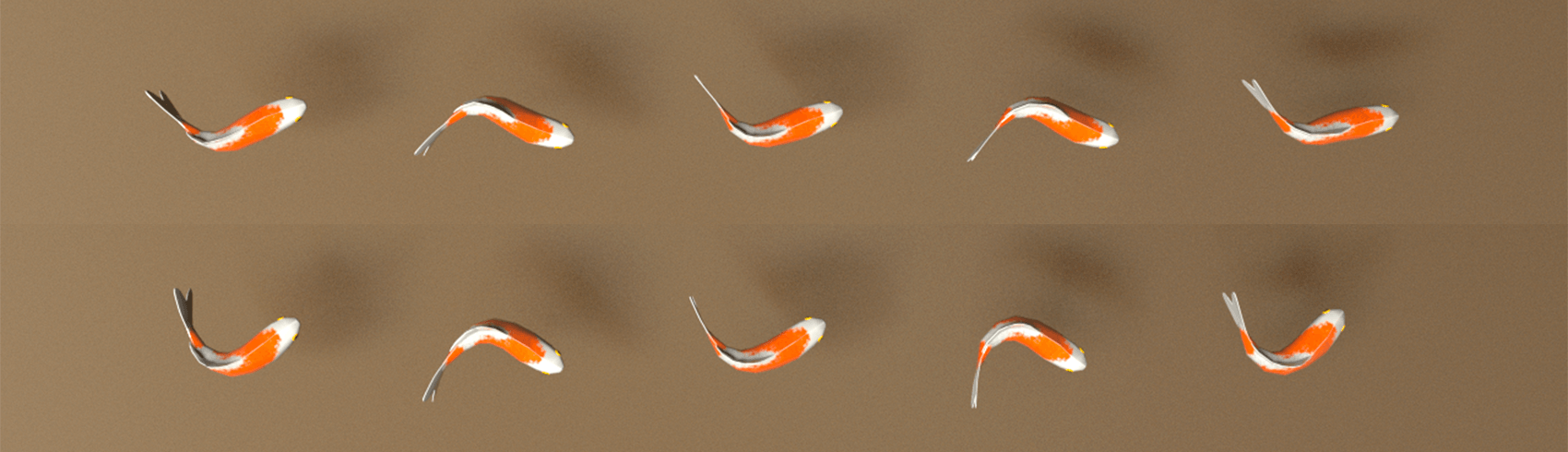} \vspace{-3mm}
	\caption{Control behaviors of a koi-like robot fish for different fluid densities. Top: $\rho=1000$; bottom: $\rho=10$.}
	\label{fig:densitycompare}
	\vspace{-18pt}
\end{figure} 

\subsection{Comparison for different fluid settings}
In contrast to the empirical model, our simulator can freely set physical parameters, leading to different control behaviors for a trained robot fish.
Here, we change the fluid density (with a higher density $\rho=1000$ and a lower density $\rho=10$, see Fig.~\ref{fig:densitycompare}) for simulation and train with the same learning parameters to achieve a cruising task.
It is observed that the robot fish in a higher density fluid seems to move with less swing amplitude, while the one in a lower density fluid has a larger swing amplitude to generate greater propulsion force, which is consistent with the physical expectations.

\subsection{Effects of reward weighting}

In this experiment we investigate effects of different reward weighting. Table~\ref{energy_table} summarizes the total energy cost by varying weights $w_v$ and  $w_e$ in the cruising task, where $w_v$ encourages fish to reach the target position while $w_e$ encourages fish to save its energy.
A good balance between energy preservation and the time to accomplish the task can be made, e.g., $w_v=1.0$ and $w_e=0.5$.

\begin{table}[h]
\vspace{-3mm}
	\caption{Impact of of weights in the reward} \vspace{-3mm}
	\label{energy_table}
	\begin{center}
	\begin{tabular}{c|c|c|c}
	\hline
	$w_v$ & $w_e$ & Total Energy Cost & Total Time (sec.)\\
	\hline
 	0.00 & 1.00 & 0.0413 & 10.0 \\
 	0.20 & 1.00 & 12.354 & 5.2 \\
 	1.00 & 1.00 & 15.046 & 4.6 \\
 	1.00 & 0.50 & 17.705 & 4.2 \\
 	1.00 & 0.00 & 40.969 & 4.2 \\
	\hline
	\end{tabular}
	\end{center}
\vspace{-15pt}
\end{table}



\section{Conclusion}
In this paper, we propose a new open-to-use simulation platform for training underwater fish-like robots.
The whole platform consists of a new modeling for fish-like underwater robot, a GPU-based non-inertial high-performance fluid-structure interaction solver (as a training environment), and reinforcement learning algorithms with both global and local policy learning.
Four different benchmark tasks were proposed and trained with our platform, with expected results.
We compared and analyzed the new training platform in terms of different results in multiple aspects to evaluate the advantages.

There are also some limitations.
First, the fish model is a reduced model that may deviate from the real robot design, and is now hence difficult to directly transfer to a real robot once learned.
Second, since we use local simulation, the platform is unable to training fish robot with a more complex external environment, e.g., a large vortex.
Finally, the grid resolution around the fish is not fine enough (otherwise, it will become very slow for training), and the accuracy is not sufficiently high.
Supporting simulation in more complex fluid environment and developing more efficient training method with higher accuracy deserve our future work.

%% file: root.bbl
\begin{thebibliography}{10}
\providecommand{\url}[1]{#1}
\csname url@samestyle\endcsname
\providecommand{\newblock}{\relax}
\providecommand{\bibinfo}[2]{#2}
\providecommand{\BIBentrySTDinterwordspacing}{\spaceskip=0pt\relax}
\providecommand{\BIBentryALTinterwordstretchfactor}{4}
\providecommand{\BIBentryALTinterwordspacing}{\spaceskip=\fontdimen2\font plus
\BIBentryALTinterwordstretchfactor\fontdimen3\font minus
  \fontdimen4\font\relax}
\providecommand{\BIBforeignlanguage}[2]{{%
\expandafter\ifx\csname l@#1\endcsname\relax
\typeout{** WARNING: IEEEtran.bst: No hyphenation pattern has been}%
\typeout{** loaded for the language `#1'. Using the pattern for}%
\typeout{** the default language instead.}%
\else
\language=\csname l@#1\endcsname
\fi
#2}}
\providecommand{\BIBdecl}{\relax}
\BIBdecl

\bibitem{Du2015}
R.~Du, Z.~Li, K.~Youcef-Toumi, and P.~V. y~Alvarado, \emph{Robot fish:
  Bio-inspired fishlike underwater robots}.\hskip 1em plus 0.5em minus
  0.4em\relax Springer, 2015.

\bibitem{Paley2021}
D.~A. Paley and N.~M. Wereley, \emph{Bioinspired Sensing, Actuation, and
  Control in Underwater Soft Robotic Systems}.\hskip 1em plus 0.5em minus
  0.4em\relax Springer, 2021.

\bibitem{Duraisamy2019}
P.~Duraisamy, R.~Sidharthan, and M.~Santhanakrishnan, ``Design, modeling, and
  control of biomimetic fish robot: A review,'' \emph{Journal of Bionic
  Engineering}, vol.~16, pp. 967--993, 11 2019.

\bibitem{Kopman2012}
V.~Kopman and M.~Porfiri, ``Design, modeling, and characterization of a
  miniature robotic fish for research and education in biomimetics and
  bioinspiration,'' \emph{IEEE/ASME Transactions on mechatronics}, vol.~18,
  no.~2, pp. 471--483, 2012.

\bibitem{Picardi2020}
G.~Picardi, M.~Chellapurath, S.~Iacoponi, S.~Stefanni, C.~Laschi, and
  M.~Calisti, ``Bioinspired underwater legged robot for seabed exploration with
  low environmental disturbance,'' \emph{Science Robotics}, vol.~5, no.~42,
  2020.

\bibitem{Berlinger2021}
F.~Berlinger, M.~Gauci, and R.~Nagpal, ``Implicit coordination for 3d
  underwater collective behaviors in a fish-inspired robot swarm,''
  \emph{Science Robotics}, vol.~6, no.~50, 2021.

\bibitem{Li2021}
G.~Li, X.~Chen, F.~Zhou, Y.~Liang, Y.~Xiao, X.~Cao, Z.~Zhang, M.~Zhang, B.~Wu,
  S.~Yin \emph{et~al.}, ``Self-powered soft robot in the mariana trench,''
  \emph{Nature}, vol. 591, no. 7848, pp. 66--71, 2021.

\bibitem{Katzschmann2018}
R.~K. Katzschmann, J.~DelPreto, R.~MacCurdy, and D.~Rus, ``Exploration of
  underwater life with an acoustically controlled soft robotic fish,''
  \emph{Science Robotics}, vol.~3, no.~16, 2018.

\bibitem{AvilaBelbutePeres2018}
F.~de~Avila Belbute-Peres, K.~Smith, K.~Allen, J.~Tenenbaum, and J.~Z. Kolter,
  ``End-to-end differentiable physics for learning and control,''
  \emph{Advances in neural information processing systems}, vol.~31, pp.
  7178--7189, 2018.

\bibitem{Lee2019}
S.~Lee, M.~Park, K.~Lee, and J.~Lee, ``Scalable muscle-actuated human
  simulation and control,'' \emph{ACM Transactions On Graphics}, vol.~38,
  no.~4, pp. 1--13, 2019.

\bibitem{James2019}
S.~James, M.~Freese, and A.~J. Davison, ``Pyrep: Bringing v-rep to deep robot
  learning,'' \emph{arXiv preprint arXiv:1906.11176}, 2019.

\bibitem{Bergamin2019}
K.~Bergamin, S.~Clavet, D.~Holden, and J.~R. Forbes, ``Drecon: data-driven
  responsive control of physics-based characters,'' \emph{ACM Transactions On
  Graphics}, vol.~38, no.~6, pp. 1--11, 2019.

\bibitem{Coumans2015}
E.~Coumans, ``Bullet physics simulation,'' in \emph{ACM SIGGRAPH 2015 Courses},
  ser. SIGGRAPH '15.\hskip 1em plus 0.5em minus 0.4em\relax New York, NY, USA:
  Association for Computing Machinery, 2015.

\bibitem{Lee2018}
J.~Lee, M.~X. Grey, S.~Ha, T.~Kunz, S.~Jain, Y.~Ye, S.~S. Srinivasa,
  M.~Stilman, and C.~K. Liu, ``Dart: Dynamic animation and robotics toolkit,''
  \emph{Journal of Open Source Software}, vol.~3, no.~22, p. 500, 2018.

\bibitem{Todorov2012}
E.~Todorov, T.~Erez, and Y.~Tassa, ``Mujoco: A physics engine for model-based
  control,'' in \emph{2012 IEEE/RSJ International Conference on Intelligent
  Robots and Systems}, 2012, pp. 5026--5033.

\bibitem{Terzopoulos1994}
D.~Terzopoulos, X.~Tu, and R.~Grzeszczuk, ``Artificial fishes: Autonomous
  locomotion, perception, behavior, and learning in a simulated physical
  world,'' \emph{Artif. Life}, vol.~1, no.~4, pp. 327--351, 1994.

\bibitem{Song2017}
J.~Song, Y.~Zhong, H.~Luo, Y.~Ding, and R.~Du, ``Hydrodynamics of larval fish
  quick turning: A computational study,'' \emph{Proceedings of the Institution
  of Mechanical Engineers, Part C: Journal of Mechanical Engineering Science},
  vol. 232, p. 095440621774327, 12 2017.

\bibitem{Song2020}
J.~Song, Y.~Zhong, R.~Du, L.~Yin, and Y.~Ding, ``Tail shapes lead to different
  propulsive mechanisms in the body/caudal fin undulation of fish,''
  \emph{Proceedings of the Institution of Mechanical Engineers, Part C: Journal
  of Mechanical Engineering Science}, vol. 235, p. 095440622096768, 11 2020.

\bibitem{Verma2018}
S.~Verma, G.~Novati, and P.~Koumoutsakos, ``Efficient collective swimming by
  harnessing vortices through deep reinforcement learning,'' \emph{Proceedings
  of the National Academy of Sciences}, vol. 115, no.~23, pp. 5849--5854, 2018.

\bibitem{Novati2017}
G.~Novati, S.~Verma, D.~Alexeev, D.~Rossinelli, W.~van Rees, and
  P.~Koumoutsakos, ``Synchronisation through learning for two self-propelled
  swimmers,'' \emph{Bioinspiration and Biomimetics}, vol.~12, p. 036001, 2017.

\bibitem{Weinstein2006}
R.~Weinstein, J.~Teran, and R.~Fedkiw, ``Dynamic simulation of articulated
  rigid bodies with contact and collision,'' \emph{IEEE Transactions on
  Visualization and Computer Graphics}, vol.~12, no.~3, pp. 365--374, 2006.

\bibitem{Li2020}
W.~Li, Y.~Chen, M.~Desbrun, C.~Zheng, and X.~Liu, ``Fast and scalable turbulent
  flow simulation with two-way coupling,'' \emph{ACM Transactions on Graphics},
  vol.~39, no.~4, Jul. 2020.

\bibitem{Chen2021gpu}
Y.~Chen, W.~Li, R.~Fan, and X.~Liu, ``Gpu optimization for high-quality kinetic
  fluid simulation,'' \emph{IEEE Transactions on Visualization and Computer
  Graphics}, 2021.

\bibitem{li2016immersed}
Z.~Li, J.~Favier, U.~D'Ortona, and S.~Poncet, ``An immersed boundary-lattice
  boltzmann method for single-and multi-component fluid flows,'' \emph{Journal
  of Computational Physics}, 2016.

\bibitem{wu2010improved}
J.~Wu and C.~Shu, ``An improved immersed boundary-lattice boltzmann method for
  simulating three-dimensional incompressible flows,'' \emph{Journal of
  Computational Physics}, 2010.

\bibitem{Raffin2019}
A.~Raffin, A.~Hill, M.~Ernestus, A.~Gleave, A.~Kanervisto, and N.~Dormann,
  ``Stable baselines3,'' \emph{GitHub repository}, 2019.

\bibitem{Paszke2019}
A.~Paszke, S.~Gross, F.~Massa, A.~Lerer, J.~Bradbury, G.~Chanan, T.~Killeen,
  Z.~Lin, N.~Gimelshein, L.~Antiga \emph{et~al.}, ``Pytorch: An imperative
  style, high-performance deep learning library,'' \emph{Advances in neural
  information processing systems}, vol.~32, pp. 8026--8037, 2019.

\bibitem{Brockman2016}
G.~Brockman, V.~Cheung, L.~Pettersson, J.~Schneider, J.~Schulman, J.~Tang, and
  W.~Zaremba, ``Openai gym,'' 2016.

\bibitem{Tan2011}
J.~Tan, Y.~Gu, G.~Turk, and C.~Liu, ``Articulated swimming creatures,''
  \emph{ACM Transactions on Graphics}, vol.~30, p.~58, 2011.

\bibitem{min2019softcon}
S.~Min, J.~Won, S.~Lee, J.~Park, and J.~Lee, ``Softcon: Simulation and control
  of soft-bodied animals with biomimetic actuators,'' \emph{ACM Transactions on
  Graphics (TOG)}, vol.~38, no.~6, pp. 1--12, 2019.

\bibitem{cieslak2019stonefish}
P.~Cie{\'s}lak, ``Stonefish: An advanced open-source simulation tool designed
  for marine robotics, with a ros interface,'' in \emph{OCEANS
  2019-Marseille}.\hskip 1em plus 0.5em minus 0.4em\relax IEEE, 2019, pp. 1--6.

\bibitem{manhaes2016uuv}
M.~M.~M. Manh{\~a}es, S.~A. Scherer, M.~Voss, L.~R. Douat, and T.~Rauschenbach,
  ``Uuv simulator: A gazebo-based package for underwater intervention and
  multi-robot simulation,'' in \emph{OCEANS 2016 MTS/IEEE Monterey}.\hskip 1em
  plus 0.5em minus 0.4em\relax IEEE, 2016, pp. 1--8.

\bibitem{fossen2011handbook}
T.~I. Fossen, \emph{Handbook of marine craft hydrodynamics and motion
  control}.\hskip 1em plus 0.5em minus 0.4em\relax John Wiley \& Sons, 2011.

\bibitem{Gan2020}
C.~Gan, J.~Schwartz, S.~Alter, M.~Schrimpf, J.~Traer, J.~D. Freitas,
  J.~Kubilius, A.~Bhandwaldar, N.~Haber, M.~Sano, K.~Kim, E.~Wang, D.~Mrowca,
  M.~Lingelbach, A.~Curtis, K.~Feigelis, D.~M. Bear, D.~Gutfreund, D.~Cox,
  J.~J. DiCarlo, J.~McDermott, J.~B. Tenenbaum, and D.~L.~K. Yamins,
  ``Threedworld: A platform for interactive multi-modal physical simulation,''
  2020.

\bibitem{Liang2018}
E.~Liang, R.~Liaw, R.~Nishihara, P.~Moritz, R.~Fox, K.~Goldberg, J.~Gonzalez,
  M.~Jordan, and I.~Stoica, ``Rllib: Abstractions for distributed reinforcement
  learning,'' in \emph{International Conference on Machine Learning}.\hskip 1em
  plus 0.5em minus 0.4em\relax PMLR, 2018, pp. 3053--3062.

\bibitem{Caspi2017}
I.~Caspi, G.~Leibovich, G.~Novik, and S.~Endrawis, ``Reinforcement learning
  coach,'' 2017.

\bibitem{godunov1959finite}
S.~Godunov and I.~Bohachevsky, ``Finite difference method for numerical
  computation of discontinuous solutions of the equations of fluid dynamics,''
  \emph{Matemati{\v{c}}eskij sbornik}, vol.~47, no.~3, pp. 271--306, 1959.

\bibitem{Rai-1991}
M.~M. Rai and P.~Moin, ``Direct simulations of turbulent flow using
  finite-difference schemes,'' \emph{Journal of computational physics},
  vol.~96, no.~1, pp. 15--53, 1991.

\bibitem{smolarkiewicz1998mpdata}
P.~K. Smolarkiewicz and L.~G. Margolin, ``Mpdata: A finite-difference solver
  for geophysical flows,'' \emph{Journal of Computational Physics}, vol. 140,
  no.~2, pp. 459--480, 1998.

\bibitem{eymard2000finite}
R.~Eymard, T.~Gallou{\"e}t, and R.~Herbin, ``Finite volume methods,''
  \emph{Handbook of numerical analysis}, vol.~7, pp. 713--1018, 2000.

\bibitem{versteeg2007introduction}
H.~K. Versteeg and W.~Malalasekera, \emph{An introduction to computational
  fluid dynamics: the finite volume method}.\hskip 1em plus 0.5em minus
  0.4em\relax Pearson education, 2007.

\bibitem{pinelli2010immersed}
A.~Pinelli, I.~Naqavi, U.~Piomelli, and J.~Favier, ``Immersed-boundary methods
  for general finite-difference and finite-volume navier--stokes solvers,''
  \emph{Journal of Computational Physics}, vol. 229, no.~24, pp. 9073--9091,
  2010.

\bibitem{wilson1983finite}
E.~L. Wilson and M.~Khalvati, ``Finite elements for the dynamic analysis of
  fluid-solid systems,'' \emph{International Journal for Numerical Methods in
  Engineering}, vol.~19, no.~11, pp. 1657--1668, 1983.

\bibitem{girault2012finite}
V.~Girault and P.-A. Raviart, \emph{Finite element methods for Navier-Stokes
  equations: theory and algorithms}.\hskip 1em plus 0.5em minus 0.4em\relax
  Springer Science \& Business Media, 2012, vol.~5.

\bibitem{elman2014finite}
H.~C. Elman, D.~J. Silvester, and A.~J. Wathen, \emph{Finite elements and fast
  iterative solvers: with applications in incompressible fluid dynamics}.\hskip
  1em plus 0.5em minus 0.4em\relax Numerical Mathematics and Scie, 2014.

\bibitem{Stam2001}
J.~Stam, ``Stable fluids,'' \emph{ACM SIGGRAPH 99}, vol. 1999, 11 2001.

\bibitem{Kim2005FlowFixer}
B.~Kim, Y.~Liu, I.~Llamas, and J.~R. Rossignac, ``Flowfixer: Using bfecc for
  fluid simulation,'' \emph{Eurographics Workshop on Natural Phenomena}, 2005.

\bibitem{Becker2007Weakly}
M.~Becker and M.~Teschner, ``Weakly compressible {SPH} for free surface
  flows,'' in \emph{Proceedings of the 2007 ACM SIGGRAPH/Eurographics symposium
  on Computer animation}, 2007.

\bibitem{Ihmsen2014}
M.~Ihmsen, J.~Orthmann, B.~Solenthaler, A.~Kolb, and M.~Teschner, ``{SPH}
  fluids in computer graphics,'' \emph{Eurographics 2014 - State of the Art
  Reports}, 2014.

\bibitem{Jiang2015Affine}
C.~Jiang, C.~Schroeder, A.~Selle, J.~Teran, and A.~Stomakhin, ``The affine
  particle-in-cell method,'' \emph{ACM Transactions on Graphics}, 2015.

\bibitem{Zehnder2018}
J.~Zehnder, R.~Narain, and B.~Thomaszewski, ``An advection-reflection solver
  for detail-preserving fluid simulation,'' \emph{ACM Transactions on
  Graphics}, 2018.

\bibitem{Qu2019}
Z.~Qu, X.~Zhang, M.~Gao, C.~Jiang, and B.~Chen, ``Efficient and conservative
  fluids using bidirectional mapping,'' \emph{ACM Transactions on Graphics},
  2019.

\bibitem{klingner2006fluid}
B.~M. Klingner, B.~E. Feldman, N.~Chentanez, and J.~F. O'brien, ``Fluid
  animation with dynamic meshes,'' in \emph{ACM SIGGRAPH 2006 Papers}, 2006,
  pp. 820--825.

\bibitem{lv2010novel}
X.~Lv, Q.~Zou, Y.~Zhao, and D.~Reeve, ``A novel coupled level set and volume of
  fluid method for sharp interface capturing on 3d tetrahedral grids,''
  \emph{Journal of Computational Physics}, vol. 229, no.~7, pp. 2573--2604,
  2010.

\bibitem{dai2005adaptive}
M.~Dai and D.~P. Schmidt, ``Adaptive tetrahedral meshing in free-surface
  flow,'' \emph{Journal of computational Physics}, vol. 208, no.~1, pp.
  228--252, 2005.

\bibitem{Liu-2012}
X.~Liu, W.-M. Pang, J.~Qin, and C.-W. Fu, ``Turbulence simulation by adaptive
  multi-relaxation lattice boltzmann modeling,'' \emph{IEEE Transactions on
  Visualization and Computer Graphics}, vol.~20, no.~02, pp. 289--302, feb
  2014.

\bibitem{Daniel-2014}
D.~Lycett-Brown, K.~H. Luo, R.~Liu, and P.~Lv, ``Binary droplet collision
  simulations by a multiphase cascaded lattice {Boltzmann} method,''
  \emph{Physics of Fluids}, vol.~26, p. 023303, 2014.

\bibitem{Rosis-2017}
A.~De~Rosis, ``Nonorthogonal central-moments-based lattice {Boltzmann} scheme
  in three dimensions,'' \emph{Physical Review E}, vol.~95, no.~1, p. 013310,
  2017.

\bibitem{Li-2018}
W.~Li, K.~Bai, and X.~Liu, ``Continuous-scale kinetic fluid simulation,''
  \emph{IEEE Transactions on Visualization and Computer Graphics}, vol.~25,
  no.~09, pp. 2694--2709, sep 2019.

\bibitem{Li-2020}
W.~Li, Y.~Chen, M.~Desbrun, C.~Zheng, and X.~Liu, ``Fast and scalable turbulent
  flow simulation with two-way coupling,'' \emph{ACM Transactions on Graphics},
  vol.~39, no.~4, Jul. 2020.

\bibitem{Nian2020}
R.~Nian, J.~Liu, and B.~Huang, ``A review on reinforcement learning:
  Introduction and applications in industrial process control,''
  \emph{Computers \& Chemical Engineering}, vol. 139, p. 106886, 2020.

\bibitem{Schulman2017a}
J.~Schulman, F.~Wolski, P.~Dhariwal, A.~Radford, and O.~Klimov, ``Proximal
  policy optimization algorithms,'' \emph{arXiv preprint arXiv:1707.06347},
  2017.

\bibitem{Mnih2016}
V.~Mnih, A.~P. Badia, M.~Mirza, A.~Graves, T.~Lillicrap, T.~Harley, D.~Silver,
  and K.~Kavukcuoglu, ``Asynchronous methods for deep reinforcement learning,''
  in \emph{International conference on machine learning}.\hskip 1em plus 0.5em
  minus 0.4em\relax PMLR, 2016, pp. 1928--1937.

\bibitem{Mnih2013}
V.~Mnih, K.~Kavukcuoglu, D.~Silver, A.~Graves, I.~Antonoglou, D.~Wierstra, and
  M.~Riedmiller, ``Playing atari with deep reinforcement learning,''
  \emph{arXiv preprint arXiv:1312.5602}, 2013.

\bibitem{Bellemare2017}
M.~G. Bellemare, W.~Dabney, and R.~Munos, ``A distributional perspective on
  reinforcement learning,'' in \emph{International Conference on Machine
  Learning}.\hskip 1em plus 0.5em minus 0.4em\relax PMLR, 2017, pp. 449--458.

\bibitem{long2018towards}
P.~Long, T.~Fan, X.~Liao, W.~Liu, H.~Zhang, and J.~Pan, ``Towards optimally
  decentralized multi-robot collision avoidance via deep reinforcement
  learning,'' in \emph{2018 IEEE International Conference on Robotics and
  Automation}.\hskip 1em plus 0.5em minus 0.4em\relax IEEE, 2018, pp.
  6252--6259.

\bibitem{li2021reinforcement}
Z.~Li, X.~Cheng, X.~B. Peng, P.~Abbeel, S.~Levine, G.~Berseth, and K.~Sreenath,
  ``Reinforcement learning for robust parameterized locomotion control of
  bipedal robots,'' \emph{arXiv preprint arXiv:2103.14295}, 2021.

\bibitem{kato2017autonomous}
Y.~Kato, K.~Kamiyama, and K.~Morioka, ``Autonomous robot navigation system with
  learning based on deep q-network and topological maps,'' in \emph{2017
  IEEE/SICE International Symposium on System Integration}.\hskip 1em plus
  0.5em minus 0.4em\relax IEEE, 2017, pp. 1040--1046.

\bibitem{xin2017application}
J.~Xin, H.~Zhao, D.~Liu, and M.~Li, ``Application of deep reinforcement
  learning in mobile robot path planning,'' in \emph{2017 Chinese Automation
  Congress}.\hskip 1em plus 0.5em minus 0.4em\relax IEEE, 2017, pp. 7112--7116.

\bibitem{chen2021non}
L.~Chen, Y.~Zhao, H.~Zhao, and B.~Zheng, ``Non-communication decentralized
  multi-robot collision avoidance in grid map workspace with double deep
  q-network,'' \emph{Sensors}, vol.~21, no.~3, p. 841, 2021.

\bibitem{Haarnoja2018}
T.~Haarnoja, A.~Zhou, P.~Abbeel, and S.~Levine, ``Soft actor-critic: Off-policy
  maximum entropy deep reinforcement learning with a stochastic actor,'' in
  \emph{International conference on machine learning}.\hskip 1em plus 0.5em
  minus 0.4em\relax PMLR, 2018, pp. 1861--1870.

\bibitem{haarnoja2018soft}
T.~Haarnoja, A.~Zhou, K.~Hartikainen, G.~Tucker, S.~Ha, J.~Tan, V.~Kumar,
  H.~Zhu, A.~Gupta, P.~Abbeel \emph{et~al.}, ``Soft actor-critic algorithms and
  applications,'' \emph{arXiv preprint arXiv:1812.05905}, 2018.

\bibitem{de2021soft}
J.~C. de~Jesus, V.~A. Kich, A.~H. Kolling, R.~B. Grando, M.~A. d. S.~L.
  Cuadros, and D.~F.~T. Gamarra, ``Soft actor-critic for navigation of mobile
  robots,'' \emph{Journal of Intelligent \& Robotic Systems}, vol. 102, no.~2,
  pp. 1--11, 2021.

\bibitem{wong2021motion}
C.-C. Wong, S.-Y. Chien, H.-M. Feng, and H.~Aoyama, ``Motion planning for
  dual-arm robot based on soft actor-critic,'' \emph{IEEE Access}, vol.~9, pp.
  26\,871--26\,885, 2021.

\bibitem{haarnoja2018learning}
T.~Haarnoja, S.~Ha, A.~Zhou, J.~Tan, G.~Tucker, and S.~Levine, ``Learning to
  walk via deep reinforcement learning,'' \emph{arXiv preprint
  arXiv:1812.11103}, 2018.

\bibitem{magnenat1988joint}
N.~Magnenat-Thalmann, R.~Laperrire, and D.~Thalmann, ``Joint-dependent local
  deformations for hand animation and object grasping,'' in \emph{In
  Proceedings on Graphics interface’88}.\hskip 1em plus 0.5em minus
  0.4em\relax Citeseer, 1988.

\bibitem{Asmuth2016}
H.~Asmuth, ``Development of overset strategies for lbm-based flow solvers,''
  Thesis, Technische Universit{\"a}t Hamburg, 2016.

\bibitem{ZhaoLi2002}
G.~Zhao-Li, Z.~Chu-Guang, and S.~Bao-Chang, ``Non-equilibrium extrapolation
  method for velocity and pressure boundary conditions in the lattice boltzmann
  method,'' \emph{Chinese Physics}, vol.~11, no.~4, pp. 366--374, 2002.

\bibitem{Peng2017}
X.~B. Peng, G.~Berseth, K.~K. Yin, and M.~Van De~Panne, ``Deeploco: Dynamic
  locomotion skills using hierarchical deep reinforcement learning,'' \emph{ACM
  Transactions on Graphics}, vol.~36, no.~4, 2017.

\bibitem{Peng2018}
X.~B. Peng, P.~Abbeel, S.~Levine, and M.~van~de Panne, ``Deepmimic:
  Example-guided deep reinforcement learning of physics-based character
  skills,'' \emph{ACM Transactions on Graphics}, vol.~37, no.~4, 2018.

\bibitem{grzeszczuk1998neuroanimator}
R.~Grzeszczuk, D.~Terzopoulos, and G.~Hinton, ``Neuroanimator: Fast neural
  network emulation and control of physics-based models,'' in \emph{Proceedings
  of the 25th annual conference on Computer graphics and interactive
  techniques}, 1998, pp. 9--20.

\bibitem{si2014realistic}
W.~Si, S.-H. Lee, E.~Sifakis, and D.~Terzopoulos, ``Realistic biomechanical
  simulation and control of human swimming,'' \emph{ACM Transactions on
  Graphics (TOG)}, vol.~34, no.~1, pp. 1--15, 2014.

\end{thebibliography}
